\documentclass[letterpaper]{article} 
\usepackage{aaai2026}  
\usepackage{times}  
\usepackage{helvet}  
\usepackage{courier}  
\usepackage[hyphens]{url}  
\usepackage{graphicx} 
\usepackage{natbib}  
\usepackage{caption} 
\usepackage{algorithm}
\usepackage{algorithmic}

\usepackage{newfloat}
\usepackage{listings}
\DeclareCaptionStyle{ruled}{labelfont=normalfont,labelsep=colon,strut=off} 
\floatstyle{ruled}
\newfloat{listing}{tb}{lst}{}
\floatname{listing}{Listing}
\usepackage{subcaption}
\usepackage{amsmath}
\usepackage{amssymb}
\usepackage{dsfont}
\usepackage{booktabs}
\usepackage{multirow}
\usepackage{pifont}

\title{AEDR: Training-Free AI-Generated Image Attribution\\ via Autoencoder Double-Reconstruction}

\author{
    Chao Wang\textsuperscript{\rm 1,}\thanks{Work partially done during the internship at iFLYTEK.}, 
    Zijin Yang\textsuperscript{\rm 1}, 
    Yaofei Wang\textsuperscript{\rm 2}, 
    Weiming Zhang\textsuperscript{\rm 1}, 
    Kejiang Chen\textsuperscript{\rm 1,}\thanks{Coresponding author.}
}
\affiliations{
    \textsuperscript{\rm 1}Anhui Province Key Laboratory of Digital Security, University of Science and Technology of China \\
    \textsuperscript{\rm 2}Hefei University of Technology\\
    chaowang0708@mail.ustc.edu.cn, bsmhmmlf@mail.ustc.edu.cn, wyf@hfut.edu.cn, zhangwm@ustc.edu.cn, chenkj@ustc.edu.cn
}

\begin{document}
\maketitle

\begin{abstract}
The rapid advancement of image-generation technologies has made it possible for anyone to create photorealistic images using generative models, raising significant security concerns. To mitigate malicious use, tracing the origin of such images is essential. Reconstruction-based attribution methods offer a promising solution, but they often suffer from reduced accuracy and high computational costs when applied to state‑of‑the‑art (SOTA) models. To address these challenges, we propose AEDR (AutoEncoder Double-Reconstruction), a novel training‑free attribution method designed for generative models with continuous autoencoders. Unlike existing reconstruction‑based approaches that rely on the value of a single reconstruction loss, AEDR performs two consecutive reconstructions using the model’s autoencoder, and adopts the ratio of these two reconstruction losses as the attribution signal. This signal is further calibrated using the image homogeneity metric to improve accuracy, which inherently cancels out absolute biases caused by image complexity, with autoencoder‑based reconstruction ensuring superior computational efficiency. Experiments on eight top latent diffusion models show that AEDR achieves 25.5\% higher attribution accuracy than existing reconstruction‑based methods, while requiring only 1\% of the computational time.
\end{abstract}

\begin{links}
    \link{Code}{https://github.com/wangchao0708/AEDR}
\end{links}

\section{Introduction}
\label{Introduction}
Latent diffusion models~\cite{SDXL, FLUX, SD, KD2.1} have emerged as the dominant paradigm in image generation due to their exceptional capabilities in producing high-resolution and photorealistic content. Their effectiveness in modeling complex data distributions has facilitated numerous applications across diverse fields, such as digital art, advertising, and virtual reality~\cite{use-2}. Recent models such as Stable Diffusion 3.5~\cite{SD3.5} and FLUX.1-dev enable rapid and controllable image synthesis, supporting a wide range of real-world applications.

While the powerful capabilities of AI‑generated images are widely embraced, concerns about potential misuse are growing~\cite{carlini2023extracting, chen2020gan, ong2021protecting, zhao2021multi}. For instance, unscrupulous vendors may repurpose outputs from third-party models as their own, falsely promoting model performance and thereby misleading consumers while undermining fair market competition. Furthermore, malicious actors can present outputs from commercial models~\cite{DALL} as original creations to gain reputational and financial advantages, flagrantly violating the intellectual property rights of model developers~\cite{IP-problem}. Therefore, reliable image origin attribution has become indispensable for accurately identifying the responsible entities behind such generated content.

\begin{figure}[t]
    \centering
    \begin{subfigure}[b]{\linewidth}
        \centering
        \includegraphics[width=0.77\linewidth]{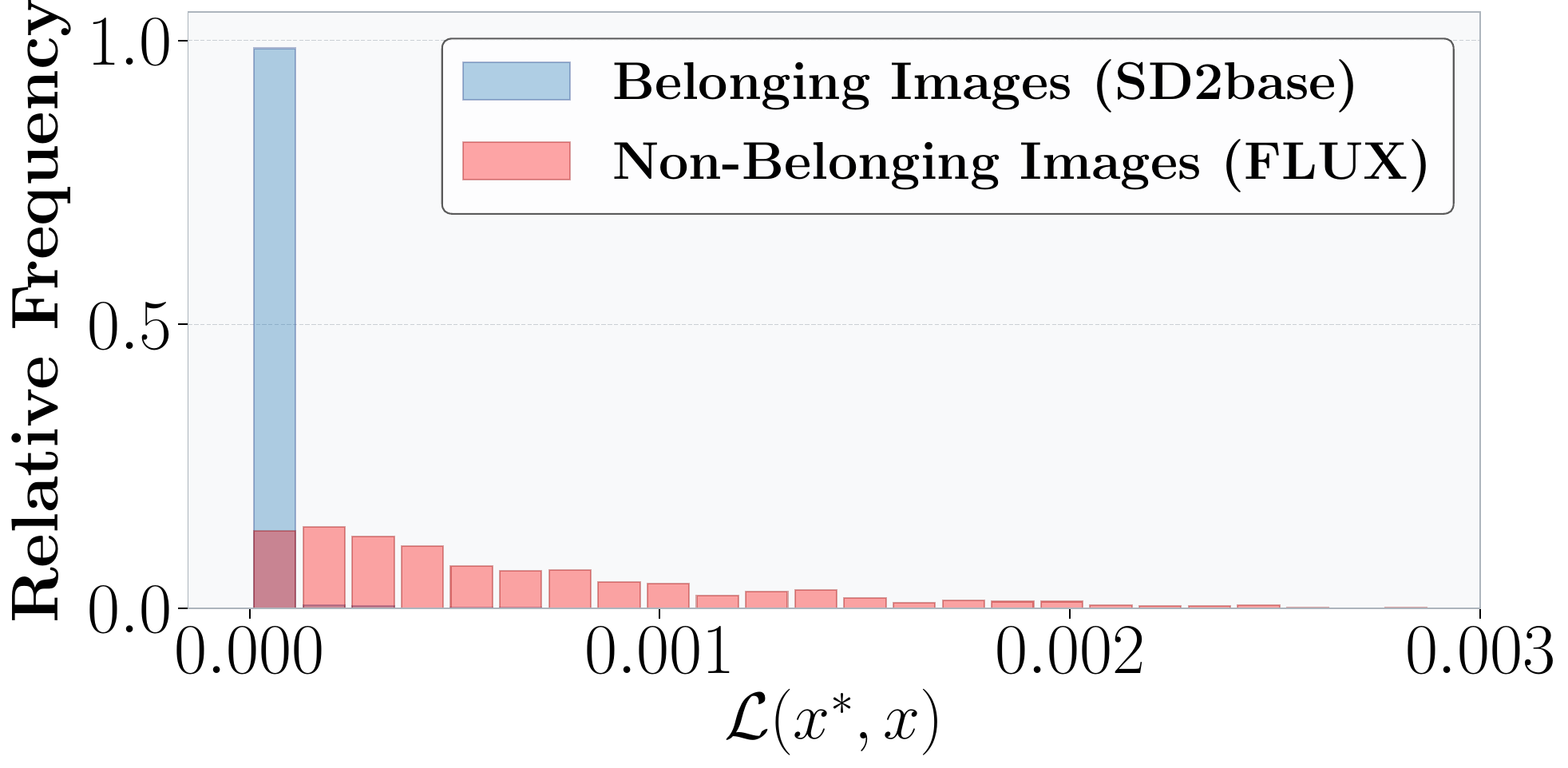}
        \caption{Gradient-based reconstruction using SD2base $(\sim10^{-4})$.}
        \label{gr-a}
    \end{subfigure}
    \begin{subfigure}[b]{\linewidth}
        \centering
        \includegraphics[width=0.77\linewidth]{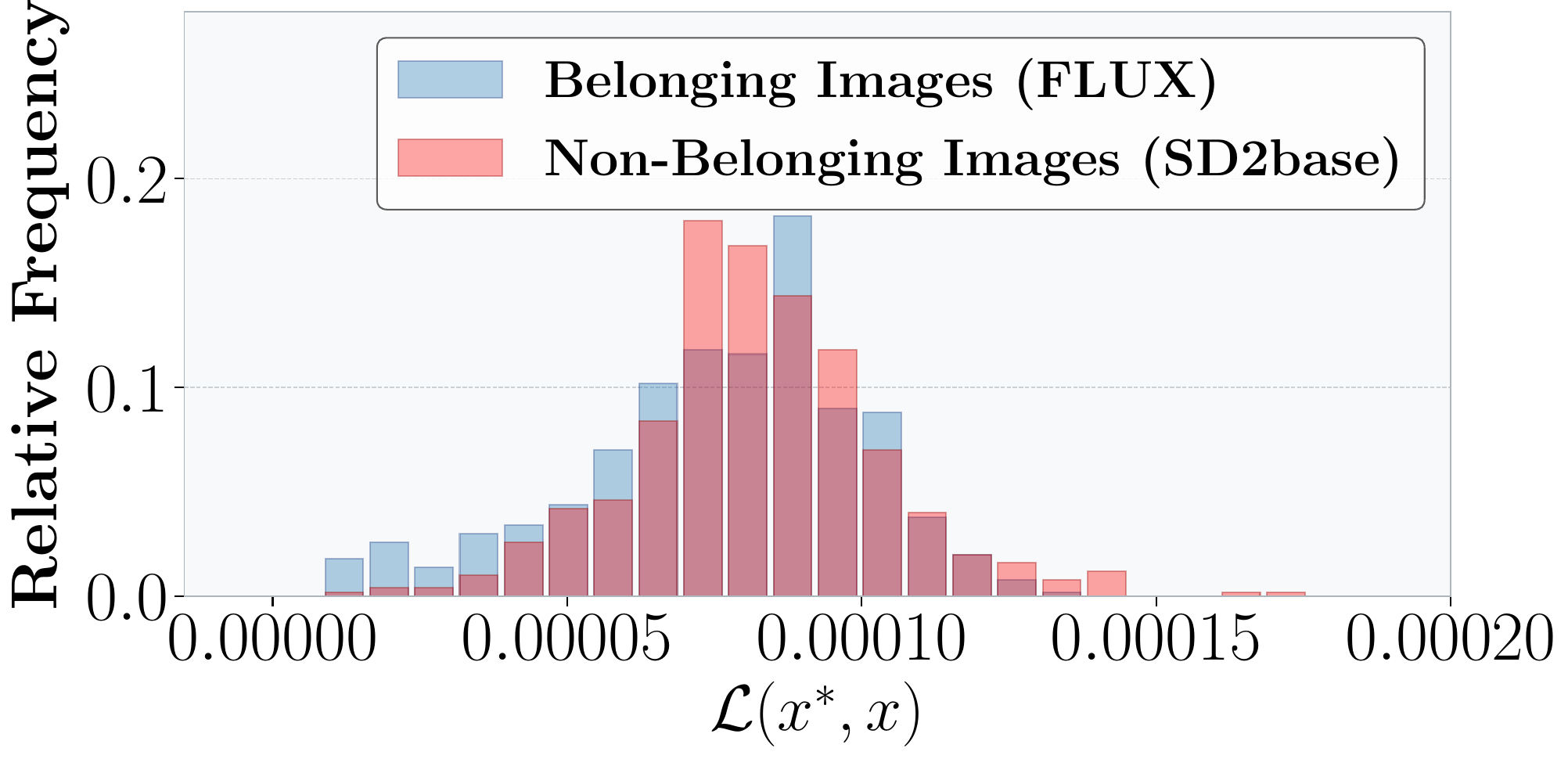}
        \caption{Gradient-based reconstruction using FLUX $(\sim10^{-5})$.}
        \label{gr-b}
    \end{subfigure}
\caption{Gradient-based reconstruction methods exhibit different loss distributions. These differences lead to attribution failures on the latest model, such as FLUX. }
\label{fig-gr-re}
\end{figure}

\begin{figure*}[t]
    \centering
    \begin{subfigure}[b]{0.33\linewidth}
        \centering
        \includegraphics[width=\linewidth]{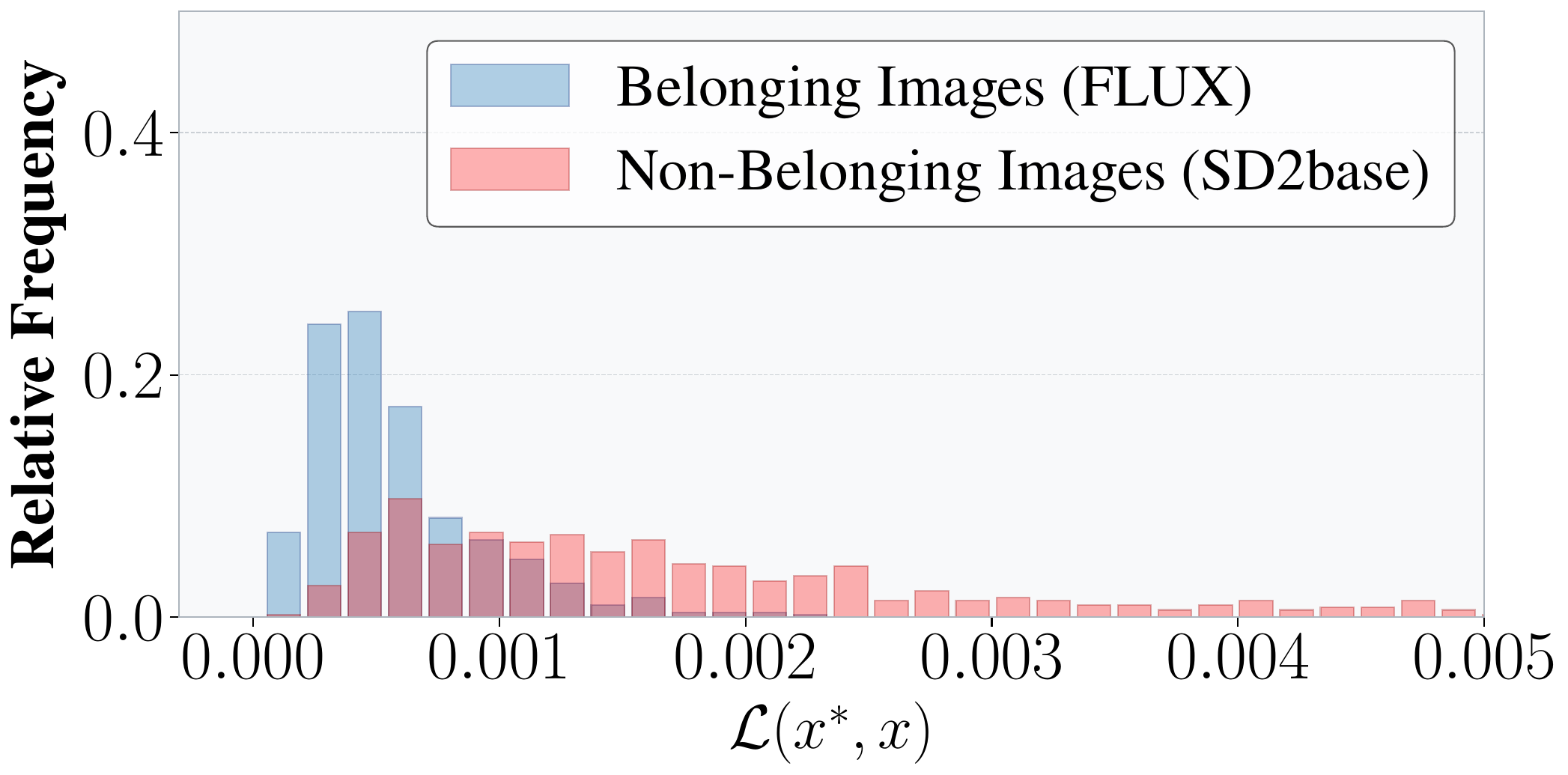}
        \caption{Distribution of the first loss.}
        \label{fig-2-1}
    \end{subfigure}
    \hfill
    \begin{subfigure}[b]{0.33\linewidth}
        \centering
        \includegraphics[width=\linewidth]{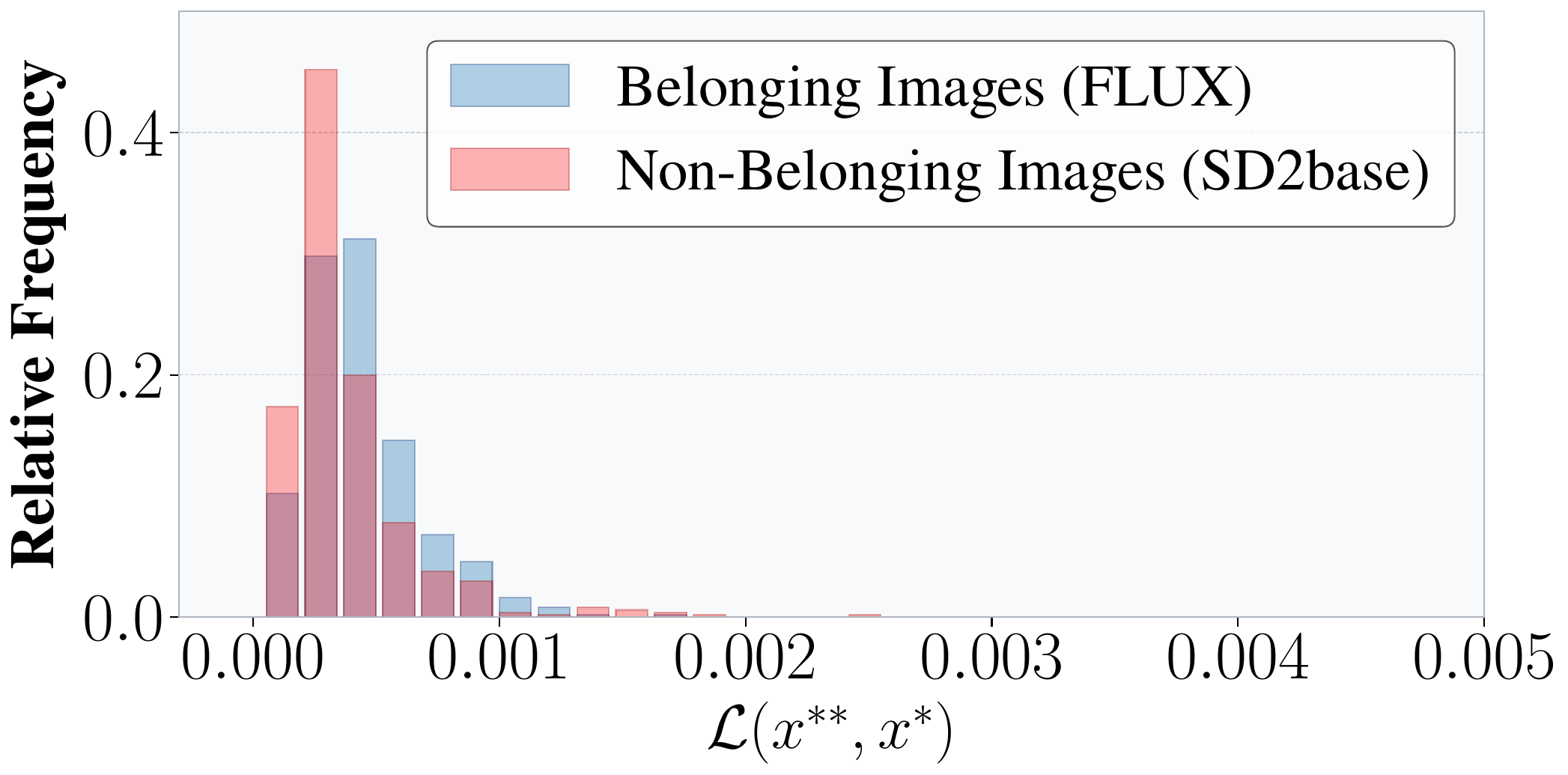}
        \caption{Distribution of the second loss.}
        \label{fig-2-2}
    \end{subfigure}
    \hfill
    \begin{subfigure}[b]{0.32\linewidth}
        \centering
        \includegraphics[width=\linewidth]{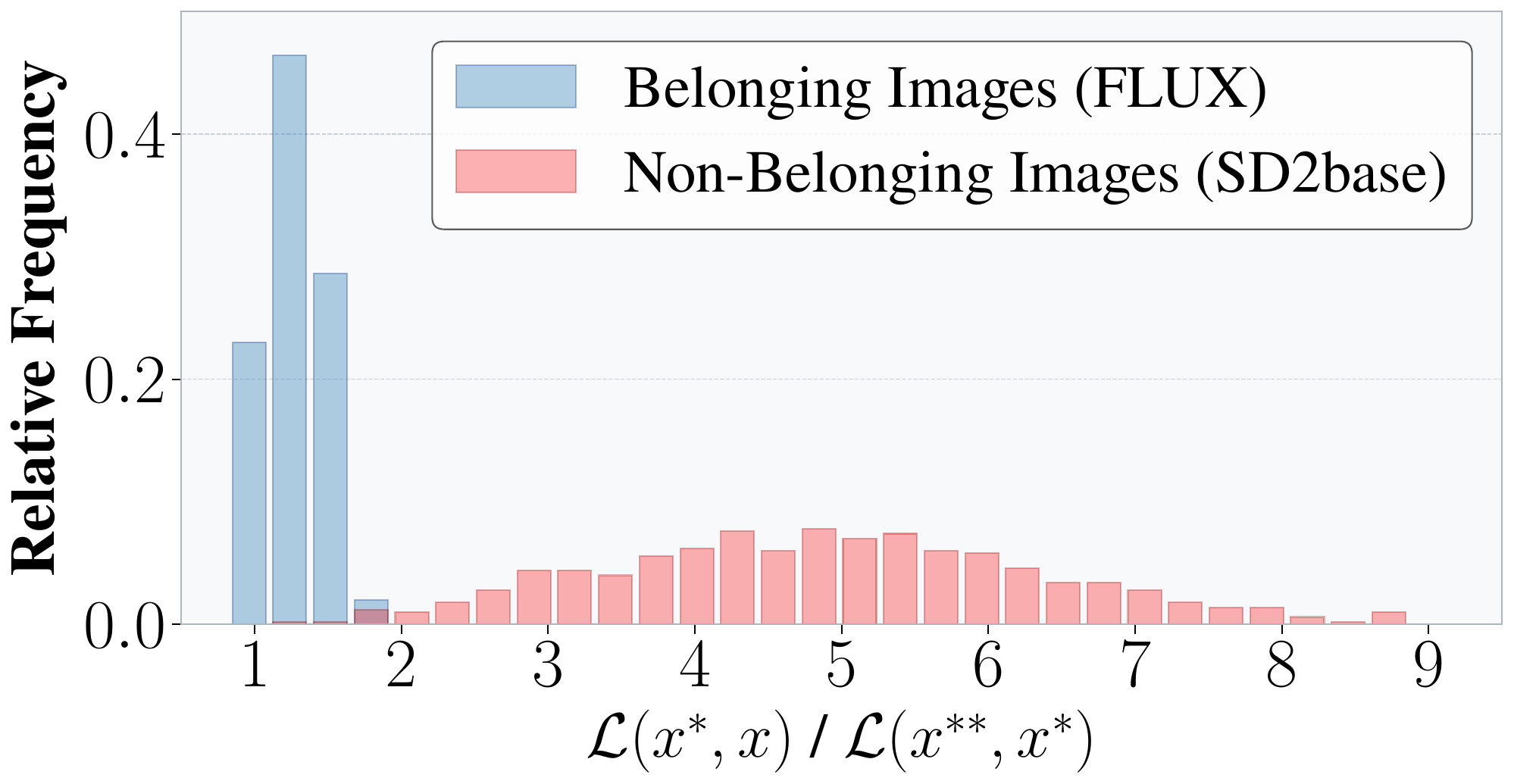}
        \caption{Distribution of the loss ratio.}
        \label{fig-2-3}
    \end{subfigure}

    \caption{Autoencoder-based double-reconstruction loss variation and loss ratio.}
    \label{Fig:ratio}
\end{figure*}

Existing image attribution methods generally fall into three primary categories. Watermark-based methods~\cite{wat-2, wat-1, Tree-Ring, Gaussian-shading, Gaussian-shading++} embed invisible or semi-visible marks during image generation, enabling attribution through watermark detection. Fingerprint-based methods~\cite{finger-2,  finger-4, finger-3} introduce model-specific signatures during training or by modifying the architecture, and rely on supervised classifiers to detect these fingerprints in generated images. However, some of these methods may degrade the visual quality of the generated images due to additional operations during training or inference. In contrast, passive detection approaches, which do not require modifications to the training process of target models, are more practical and acceptable in real-world applications. Among these, reconstruction-based methods~\cite{RONAN, LatentTracer} have achieved excellent performance. They use gradient information from the generative model to reconstruct the input image, with the reconstruction loss between the original and reconstructed images serving as the attribution signal. These methods do not modify model parameters or the generation pipeline, and thus preserve generative performance, making them one of the most promising approaches for image attribution.  

Existing gradient-based reconstruction methods typically follow this observation: images that belong to the target model generally exhibit lower reconstruction losses than non-belonging images (see Figure~\ref{gr-a}). Thus, the two types of images can be distinguished based on a single reconstruction loss value. However, with increasingly powerful generative models such as FLUX~\cite{FLUX}, these methods~\cite{RONAN, LatentTracer} tend to produce extremely low reconstruction losses for both belonging and non-belonging images, causing significant overlap in the loss distributions and thereby reducing attribution accuracy (see Figure~\ref{gr-b}). Moreover, the gradient-guided reconstruction process is computationally expensive and complex, making these methods impractical for real-world applications.

To address these limitations, we prefer to adopt autoencoder-based reconstruction as the foundation for an efficient attribution. However, using reconstruction loss directly is susceptible to the inherent complexity of the input image~\cite{HFI, AEroblade}: Images with simple textures naturally have lower reconstruction loss, while those with complex textures correspondingly have higher reconstruction loss (Further details are provided in Sections 2 and 3 of the Technical Appendix). To mitigate this bias introduced by image complexity, we propose an attribution method based on the loss ratio across double-reconstructions by the autoencoder. The key insight is that the autoencoder has learned to capture representative features of samples from the training distribution. As illustrated in Figure~\ref{fig-2-1} and Figure~\ref{fig-2-2}, we observe that when the autoencoder reconstructs a belonging image twice, the resulting reconstruction losses are nearly identical, since the image lies within the model's distribution. In contrast, for the non-belonging image, which initially lies outside the model distribution, the first reconstruction effectively projects it into the distribution, leading to a noticeably lower loss in the second reconstruction. Given the pronounced disparity, if we compute the ratio of the first to the second reconstruction loss, the value tends to be close to 1 for belonging images, but significantly greater than 1 for non-belonging images (see Figure~\ref{fig-2-3}).

Motivated by this insight, we propose an image attribution method based on \textbf{A}uto\textbf{E}ncoder \textbf{D}ouble-\textbf{R}econstruction, namely \textbf{AEDR}. Unlike existing gradient-guided approaches that utilize single reconstruction loss, AEDR employs the loss ratio from double reconstructions, calibrated by an image homogeneity metric. This approach mitigates discrepancies arising from image texture complexity. Additionally, we implement a Kernel Density Estimation method~\cite{KDE} for adaptive threshold selection, which assumes no prior knowledge of underlying data distributions, thereby enhancing the adaptability across diverse generative models.Our contributions are summarized as follows:
\begin{itemize}
    \item We reveal that existing reconstruction-based image attribution methods, which rely on the single reconstruction loss, face significant challenges when applied to SOTA generative models. We identify a key discrepancy: under autoencoder-based double-reconstruction, belonging images show significantly higher latent feature consistency across reconstructions than non-belonging ones.

    \item We propose AEDR, an training‑free passive attribution method that leverages the ratio of losses from two successive reconstructions and calibrates it based on image homogeneity. AEDR performs direct reconstruction using the model’s autoencoder, eliminating the need for gradient-based optimization or additional training.

    \item Extensive experiments on eight SOTA or widely used generative models demonstrate that AEDR improves the attribution accuracy by an average of 25.5\% and reduces the attribution time to just 1\% of the best baseline.
\end{itemize}


\section{Related Work}
\label{Related Work}

\paragraph{Image Generative Models.}
Early frameworks like Generative Adversarial Networks (GANs)~\cite{GAN}, Variational Autoencoders (VAEs)~\cite{VAE}, and autoregressive methods such as PixelCNN~\cite{van2016pixel} laid the groundwork for realistic image synthesis. However, diffusion models have recently become the leading approach. Denoising Diffusion Probabilistic Models (DDPMs)~\cite{DDPM} achieve remarkable image fidelity by iteratively adding and reversing Gaussian noise, using simplified objectives and U-Net architectures. Building on this, Latent Diffusion Models (LDMs) improved efficiency by processing diffusion within a compact latent space, enabling rapid, high-resolution image generation with significantly lower computational requirements. 

\paragraph{Detection of AI-Generated Images.}
As generative images grow increasingly photorealistic, the need for reliable detection is increasing~\cite{chen2023pathway}. Early work by Marra combined CycleGAN with steganalysis to identify GAN-generated images, laying a foundation for this field. Most existing methods frame the task as a binary classification problem, distinguishing between real and generated images. These approaches often rely on discriminative texture features~\cite{tex-2, tex-4, tex-7} or frequency domain signals~\cite{fre-4, fre-5, fre-7} as core detection cues. Recent approaches such as AEROBLADE~\cite{AEroblade} and HFI~\cite{HFI} adopt reconstruction losses via autoencoders and use LPIPS as a detection metric. Although these methods demonstrate promising performance in generative image detection, they do not address the more challenging task of origin attribution.

\paragraph{Origin Attribution of Generated Images.}
Current methods for attributing the origin of generated images can be broadly classified into three types: watermark-based methods~\cite{wat-1, Tree-Ring, Gaussian-shading, Gaussian-shading++}, fingerprint-based methods~\cite{finger-2, finger-1, finger-4, finger-3}, and reconstruction-based methods~\cite{RONAN, LatentTracer}. Watermark-based methods embed model-specific information into the generated image and retrieve this information during detection to identify the origin. Fingerprint-based methods inject unique patterns into the model during training and rely on dedicated classifiers for attribution. However, both active attribution approaches require additional intervention during image generation or model training and may compromise the quality of the generated images. In contrast, reconstruction-based passive methods enable origin attribution without the need for any extra operations. For example, both RONAN~\cite{RONAN} and LatentTracer~\cite{LatentTracer} adopt gradient-based reconstruction approaches, using the loss between the reconstructed and original images as the attribution signal. 

\begin{figure*}[t]
    \centering
    \includegraphics[width=0.87\linewidth]{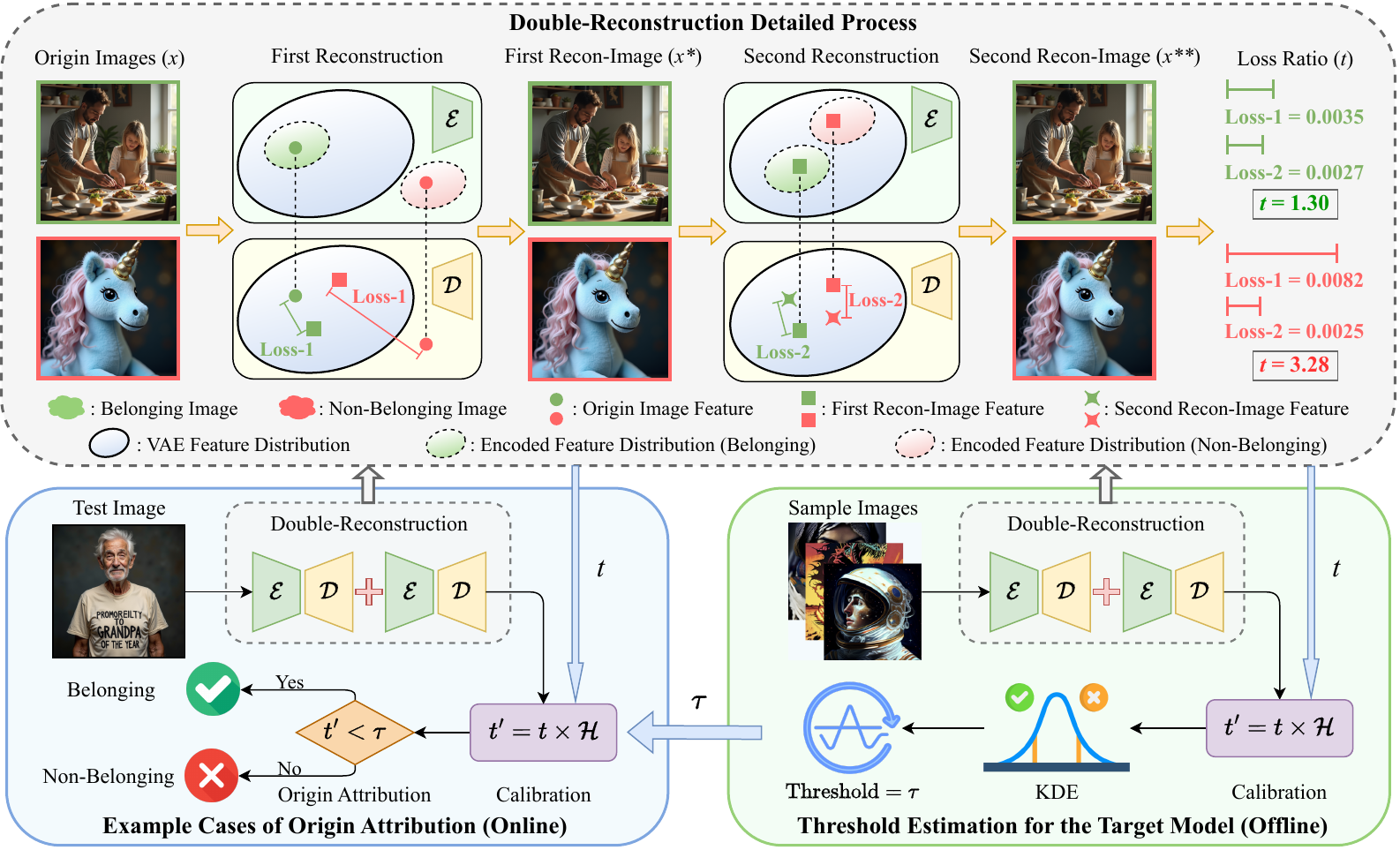}
    \caption{The framework of AEDR. Our method consists of three key modules: double-reconstruction based on autoencoder (Section~\ref{method-1}), calibration mechanism (Section~\ref{method-2}), and threshold determination via kernel density estimation (Section~\ref{method-3}). AEDR employs the ratio of double-reconstruction losses, calibrated by an image homogeneity metric, as the attribution signal. The decision threshold is determined via kernel density estimation and applied for origin attribution.}
    \label{fig:Framework}
\end{figure*}

\section{Method}
\label{section 4}

This section provides a detailed description of AEDR, the passive attribution method proposed in this paper. The process begins with two independent reconstructions of the input image using an autoencoder. The ratio of their reconstruction losses is then computed. To improve robustness, this ratio is further calibrated using an image homogeneity metric, yielding the final attribution signal. The attribution outcome is determined by comparing this signal against a predefined threshold. The overall structure of the AEDR framework is depicted in Figure~\ref{fig:Framework}.

\subsection{Problem Statement and Threat Models}

We begin by formalizing the key concepts related to the attribution task, followed by the definition of the attribution goal and the specification of the threat model.

\textbf{Belonging vs. Non-Belonging Image.}
Given an image generative model \( \mathcal{M}: \mathcal{I} \rightarrow \mathcal{X}_\mathcal{M} \) , where \( \mathcal{I} \) is the input space and \( \mathcal{X}_\mathcal{M} \subset \mathcal{X} \) denotes the subset of images produced by \( \mathcal{M} \), with \( \mathcal{X} \) representing the full image space. A test image \(x\) is classified as a belonging image if and only if \(x \in \mathcal{X}_{\mathcal{M}}\); if \(x \notin \mathcal{X}_{\mathcal{M}}\), it is classified as a non-belonging image.

\textbf{Attribution Goal.}
The goal of the auditor is to determine whether a given image \( x \) was generated by the target model \( \mathcal{M} \), without modifying the training or inference process of \( \mathcal{M} \), and without applying any post-processing to its outputs. This is formulated as a binary function \( F: \{\mathcal{M}, x\} \rightarrow \{0, 1\} \), where \( F(\mathcal{M}, x) = 0 \) indicates that \( x \) is attributed to \( \mathcal{M} \), and \( F(\mathcal{M}, x) = 1 \) indicates a non-belonging image.

\textbf{Threat Model.}
We assume a \textit{white-box} setting in which the auditor has access only to the autoencoder \(\mathcal{R}\) associated with the model \(\mathcal{M}\). Specifically, the auditor can query both the encoder and decoder for image reconstruction, but does not have access to model gradients, training data, or internal parameters of \(\mathcal{M}\). This is in contrast to prior work, which require full white-box access, including gradient computation. 
By relying solely on reconstruction queries, AEDR is more practical for deployment in restricted-access scenarios.

\subsection{Double-Reconstruction via Autoencoder}
\label{method-1}

The specific workflow is as follows: Given the autoencoder \( \mathcal{R} \) of the target model \( \mathcal{M} \) and a test image \( x \), we perform two successive reconstructions. The two reconstruction steps are respectively defined in Equations~\eqref{eq:recon1} and~\eqref{eq:recon2}.
\begin{equation}
x^* = \mathcal{R}(x) = \mathcal{D}(\mathcal{E}(x)), \quad \mathcal{L}_1 = \mathcal{L}(x^*, x) .
\label{eq:recon1}
\end{equation}
\begin{equation}
x^{**} = \mathcal{R}(x^*) = \mathcal{D}(\mathcal{E}(x^*)), \quad \mathcal{L}_2 = \mathcal{L}(x^{**}, x^*) .
\label{eq:recon2}
\end{equation}
Here, \( x^* \) and \( x^{**} \) denote the reconstructed images obtained from the first and second passes through the autoencoder \( \mathcal{R} \). The corresponding reconstruction losses are denoted by \( \mathcal{L}_1 \) and \( \mathcal{L}_2 \), both computed using mean squared error (MSE). The advantage of using MSE is demonstrated in Experiment~\ref{Experiment-3}. AEDR focuses on analyzing the trend of variation in reconstruction losses for the test image \( x \) during the double-reconstruction process. Specifically, we define an uncalibrated attribution signal \( t \) as the ratio between the first and second reconstruction losses, formally given by: 
\begin{equation}
t = \frac{\mathcal{L}_1}{\mathcal{L}_2} = \frac{\mathcal{L}(x^*, x)}{\mathcal{L}(x^{**}, x^*)} .
\end{equation}

\subsection{Calibration Mechanism}
\label{method-2}

The inherent visual complexity of an image can significantly affect its reconstruction loss. Images with homogeneous backgrounds and low texture variation typically exhibit minimal differences in double-reconstruction losses. In contrast, images containing complex textures or dynamic backgrounds often lead to substantial fluctuations in these losses (error examples see Technical Appendix, Section 6). To address this intrinsic bias, we introduce a homogeneity-aware loss calibration mechanism, defined as:
\begin{equation}
\mathcal{H} = \sum_{i=0}^{\ell-1} \sum_{j=0}^{\ell-1} \frac{P(i, j)}{1 + |i - j|} ,
\end{equation}
where \(P(i, j)\) denotes the co-occurrence probability of grayscale levels \(i\) and \(j\) at a specified orientation and offset. The parameter \(\ell\) represents the total number of grayscale levels (default value of 32 to ensure computational efficiency), and \(|i - j|\) indicates the absolute difference in intensity between the corresponding pixel pairs. Based on this, the calibrated attribution signal is defined as follows:
\begin{equation}
t^{\prime}=t\times \mathcal{H}=\frac{\mathcal{L}_1 \times \mathcal{H}}{\mathcal{L}_2} .
\end{equation}
The calibrated metric \(t^{\prime}\) effectively mitigates the intrinsic complexity bias in attribution assessment, thereby enhancing the overall attribution accuracy of AEDR.

\subsection{Threshold Determination}
\label{method-3}

Experimental results show that the calibrated attribution signal \( t^{\prime} \) does not follow a consistent or well-defined probability distribution across different models (see Technical Appendix, Section 5), and may contain a small number of outliers. To address this, we adopt kernel density estimation (KDE)~\cite{KDE}, a non-parametric method that makes no assumptions about the underlying distribution and is inherently robust to outliers. The attribution threshold is then derived from the cumulative distribution function (CDF) estimated via KDE, as defined by:
\begin{equation}
\tau = \inf\Bigl\{u \;\Bigm|\;\int_{-\infty}^{u}\frac{1}{N\,h}\sum_{i=1}^N K\!\Bigl(\frac{y - t^{\prime}_i}{h}\Bigr)\,\mathrm{d}y \;\ge\;1-\alpha\Bigr\} ,
\end{equation}
where \( N \) (set to 500 in our experiments) denotes the number of samples. Specifically, we estimate the distribution using 500 calibrated attribution signals obtained from belonging images. The parameter \( h \) represents the bandwidth of the kernel density estimation~\cite{KDEevaluation}, and \( K \) is the kernel function, for which a Gaussian kernel is used. The variable \( t_i' \) denotes the calibrated attribution signal of the \( i \)-th image. The quantity \( 1 - \alpha \) specifies the target cumulative probability for threshold selection. If the calibrated attribution signal \( t^{\prime} < \tau \), the \( x \) is classified as a belonging image; otherwise, it is considered a non-belonging image. 

\begin{table*}[t!]
\centering
\small
\setlength{\tabcolsep}{1mm}
\begin{tabular}{ccccccccccccccccccc}
\toprule
\multirow{2.5}{*}{$\mathcal{M}_1$} & \multirow{2.5}{*}{$\mathcal{M}_2$} & \multicolumn{5}{c}{RONAN} &  & \multicolumn{5}{c}{LatentTracer} &  & \multicolumn{5}{c}{AEDR} \\
\cmidrule(r){3-7} \cmidrule(lr){9-13} \cmidrule(l){15-19}
 &  & TP & FP & FN & TN & Acc &  & TP & FP & FN & TN & Acc &  & TP & FP & FN & TN & Acc \\
\midrule
\multirow{7}{*}{SD1.5} & SD2.1 & 476 & 475 & 24 & 25 & 50.1\% &  & 475 & 62 & 25 & 438 & 91.3\% &  & 493 & 13 & 7 & 487 & \textbf{98.0\%} \\
 & SD2base & 476 & 492 & 24 & 8 & 48.4\% &  & 475 & 415 & 25 & 459 & 93.4\% &  & 493 & 13 & 7 & 497 & \textbf{98.0\%} \\
 & SD3.5 & 476 & 460 & 24 & 40 & 51.6\% &  & 475 & 43 & 25 & 457 & 93.2\% &  & 493 & 3 & 7 & 497 & \textbf{99.0\%} \\
 & SDXL & 476 & 491 & 24 & 9 & 48.5\% &  & 475 & 68 & 25 & 432 & 90.7\% &  & 493 & 10 & 7 & 490 & \textbf{98.3\%} \\
 & FLUX & 476 & 484 & 24 & 16 & 49.2\% &  & 475 & 78 & 25 & 422 & 89.7\% &  & 493 & 12 & 7 & 488 & \textbf{98.1\%} \\
 & VQDM & 476 & 499 & 24 & 1 & 47.7\% &  & 475 & 37 & 25 & 463 & 93.8\% &  & 493 & 5 & 7 & 495 & \textbf{98.8\%} \\
 & KD2.1 & 476 & 471 & 24 & 29 & 50.5\% &  & 475 & 77 & 25 & 423 & 89.8\% &  & 493 & 4 & 7 & 496 & \textbf{98.9\%} \\
 \midrule
\multirow{6}{*}{SD2.1} & SD1.5 & 477 & 487 & 23 & 13 & 49.0\% &  & 476 & 244 & 24 & 256 & 73.2\% &  & 499 & 2 & 1 & 498 & \textbf{99.7\%} \\
 & SD3.5 & 477 & 408 & 23 & 92 & 56.9\% &  & 476 & 288 & 24 & 212 & 68.8\% &  & 499 & 8 & 1 & 492 & \textbf{99.1\%} \\
 & SDXL & 477 & 500 & 23 & 0 & 47.7\% &  & 476 & 443 & 24 & 57 & 53.3\% &  & 499 & 1 & 1 & 499 & \textbf{99.8\%} \\
 & FLUX & 477 & 480 & 23 & 20 & 49.7\% &  & 476 & 409 & 24 & 91 & 56.7\% &  & 499 & 4 & 1 & 496 & \textbf{99.5\%} \\
 & VQDM & 477 & 499 & 23 & 1 & 47.8\% &  & 476 & 340 & 24 & 160 & 63.6\% &  & 499 & 0 & 1 & 500 & \textbf{99.9\%} \\
 & KD2.1 & 477 & 495 & 23 & 5 & 48.2\% &  & 476 & 372 & 24 & 128 & 60.4\% &  & 499 & 1 & 1 & 499 & \textbf{99.8\%} \\
 \midrule
\multirow{6}{*}{SD2base} & SD1.5 & 475 & 409 & 25 & 91 & 56.6\% &  & 480 & 1 & 20 & 499 & 97.9\% &  & 499 & 11 & 1 & 489 & \textbf{98.8\%} \\
 & SD3.5 & 475 & 387 & 25 & 113 & 58.8\% &  & 480 & 4 & 20 & 492 & 97.2\% &  & 499 & 14 & 1 & 486 & \textbf{98.5\%} \\
 & SDXL & 475 & 476 & 25 & 24 & 49.9\% &  & 480 & 9 & 20 & 491 & 97.1\% &  & 499 & 6 & 1 & 494 & \textbf{99.3\%} \\
 & FLUX & 475 & 460 & 25 & 40 & 51.5\% &  & 480 & 28 & 20 & 472 & 95.2\% &  & 499 & 17 & 1 & 483 & \textbf{98.2\%} \\
 & VQDM & 475 & 497 & 25 & 3 & 47.8\% &  & 480 & 5 & 20 & 495 & 97.5\% &  & 499 & 1 & 1 & 499 & \textbf{99.8\%} \\
 & KD2.1 & 475 & 437 & 25 & 63 & 53.8\% &  & 480 & 30 & 20 & 470 & 95.0\% &  & 499 & 5 & 1 & 495 & \textbf{99.4\%} \\
 \midrule
\multirow{7}{*}{SD3.5} & SD1.5 & 479 & 500 & 21 & 0 & 47.9\% &  & 478 & 459 & 22 & 41 & 51.9\% &  & 490 & 7 & 10 & 493 & \textbf{98.3\%} \\
 & SD2.1 & 479 & 499 & 21 & 1 & 48.0\% &  & 478 & 486 & 22 & 14 & 49.2\% &  & 490 & 11 & 10 & 489 & \textbf{97.9\%} \\
 & SD2base & 479 & 498 & 21 & 2 & 48.1\% &  & 478 & 489 & 22 & 11 & 48.9\% &  & 490 & 7 & 10 & 493 & \textbf{98.3\%} \\
 & SDXL & 479 & 499 & 21 & 1 & 48.0\% &  & 478 & 500 & 22 & 0 & 47.8\% &  & 490 & 42 & 10 & 458 & \textbf{94.8\%} \\
 & FLUX & 479 & 479 & 21 & 21 & 50.0\% &  & 478 & 500 & 22 & 0 & 47.8\% &  & 490 & 78 & 10 & 422 & \textbf{91.2\%} \\
 & VQDM & 479 & 499 & 21 & 1 & 48.0\% &  & 478 & 492 & 22 & 8 & 48.6\% &  & 490 & 2 & 10 & 498 & \textbf{98.8\%} \\
 & KD2.1 & 479 & 467 & 21 & 33 & 51.2\% &  & 478 & 482 & 22 & 18 & 49.6\% &  & 490 & 22 & 10 & 478 & \textbf{96.8\%} \\
 \midrule
\multirow{7}{*}{SDXL} & SD1.5 & 474 & 468 & 26 & 32 & 50.6\% &  & 478 & 227 & 22 & 273 & 75.1\% &  & 500 & 5 & 0 & 495 & \textbf{99.5\%} \\
 & SD2.1 & 474 & 469 & 26 & 31 & 50.5\% &  & 478 & 340 & 22 & 160 & 63.8\% &  & 500 & 3 & 0 & 497 & \textbf{99.7\%} \\
 & SD2base & 474 & 494 & 26 & 6 & 48.0\% &  & 478 & 295 & 22 & 205 & 68.3\% &  & 500 & 6 & 0 & 494 & \textbf{99.4\%} \\
 & SD3.5 & 474 & 340 & 26 & 160 & 63.4\% &  & 478 & 292 & 22 & 208 & 68.6\% &  & 500 & 5 & 0 & 495 & \textbf{99.5\%} \\
 & FLUX & 474 & 442 & 26 & 58 & 53.2\% &  & 478 & 421 & 22 & 79 & 55.7\% &  & 500 & 15 & 0 & 485 & \textbf{98.5\%} \\
 & VQDM & 474 & 498 & 26 & 2 & 47.6\% &  & 478 & 349 & 22 & 151 & 62.9\% &  & 500 & 1 & 0 & 499 & \textbf{99.9\%} \\
 & KD2.1 & 474 & 484 & 26 & 16 & 49.0\% &  & 478 & 364 & 22 & 136 & 61.4\% &  & 500 & 2 & 0 & 498 & \textbf{99.8\%} \\
 \midrule
\multirow{7}{*}{FLUX} & SD1.5 & 474 & 500 & 26 & 0 & 47.4\% &  & 478 & 407 & 22 & 93 & 57.1\% &  & 495 & 2 & 5 & 498 & \textbf{99.3\%} \\
 & SD2.1 & 474 & 499 & 26 & 1 & 47.5\% &  & 478 & 434 & 22 & 66 & 54.4\% &  & 495 & 9 & 5 & 491 & \textbf{98.6\%} \\
 & SD2base & 474 & 497 & 26 & 3 & 47.7\% &  & 478 & 460 & 22 & 40 & 51.8\% &  & 495 & 6 & 5 & 494 & \textbf{98.9\%} \\
 & SD3.5 & 474 & 472 & 26 & 28 & 50.2\% &  & 478 & 465 & 22 & 35 & 51.3\% &  & 495 & 41 & 5 & 459 & \textbf{95.4\%} \\
 & SDXL & 474 & 499 & 26 & 1 & 47.5\% &  & 478 & 476 & 22 & 24 & 50.2\% &  & 495 & 117 & 5 & 383 & \textbf{87.8\%} \\
 & VQDM & 474 & 305 & 26 & 195 & 66.9\% &  & 478 & 462 & 22 & 38 & 51.6\% &  & 495 & 1 & 5 & 499 & \textbf{99.4\%} \\
 & KD2.1 & 474 & 469 & 26 & 31 & 50.5\% &  & 478 & 475 & 22 & 25 & 50.3\% &  & 495 & 27 & 5 & 473 & \textbf{96.8\%} \\
 \midrule
\multirow{7}{*}{VQDM} & SD1.5 & 476 & 487 & 24 & 13 & 48.9\% &  & 477 & 390 & 23 & 110 & 58.7\% &  & 469 & 105 & 31 & 395 & \textbf{86.4\%} \\
 & SD2.1 & 476 & 472 & 24 & 28 & 50.4\% &  & 477 & 415 & 23 & 85 & 56.2\% &  & 469 & 94 & 31 & 406 & \textbf{87.5\%} \\
 & SD2base & 476 & 493 & 24 & 7 & 48.3\% &  & 477 & 402 & 23 & 98 & 57.5\% &  & 469 & 83 & 31 & 417 & \textbf{88.6\%} \\
 & SD3.5 & 476 & 463 & 24 & 37 & 51.3\% &  & 477 & 446 & 23 & 54 & 53.1\% &  & 469 & 31 & 31 & 469 & \textbf{93.8\%} \\
 & SDXL & 476 & 490 & 24 & 10 & 48.6\% &  & 477 & 473 & 23 & 27 & 50.4\% &  & 469 & 50 & 31 & 450 & \textbf{91.9\%} \\
 & FLUX & 476 & 473 & 24 & 27 & 50.3\% &  & 477 & 481 & 23 & 19 & 49.6\% &  & 469 & 37 & 31 & 463 & \textbf{93.2\%} \\
 & KD2.1 & 476 & 468 & 24 & 32 & 50.8\% &  & 477 & 421 & 23 & 79 & 55.6\% &  & 469 & 40 & 31 & 460 & \textbf{92.9\%} \\
 \midrule
\multirow{7}{*}{KD2.1} & SD1.5 & 481 & 483 & 19 & 17 & 49.8\% &  & 479 & 14 & 21 & 486 & \textbf{96.5\%} &  & 472 & 157 & 28 & 343 & 81.5\% \\
 & SD2.1 & 481 & 487 & 19 & 13 & 49.4\% &  & 479 & 40 & 21 & 460 & \textbf{93.9\%} &  & 472 & 215 & 28 & 285 & 75.7\% \\
 & SD2base & 481 & 495 & 19 & 5 & 48.6\% &  & 479 & 24 & 21 & 476 & \textbf{95.5\%} &  & 472 & 155 & 28 & 345 & 81.7\% \\
 & SD3.5 & 481 & 497 & 19 & 3 & 48.4\% &  & 479 & 36 & 21 & 464 & \textbf{94.3\%} &  & 472 & 146 & 28 & 354 & 82.6\% \\
 & SDXL & 481 & 493 & 19 & 7 & 48.8\% &  & 479 & 74 & 21 & 426 & \textbf{90.5\%} &  & 472 & 188 & 28 & 312 & 78.4\% \\
 & FLUX & 481 & 493 & 19 & 7 & 48.8\% &  & 479 & 82 & 21 & 418 & \textbf{89.7\%} &  & 472 & 159 & 28 & 341 & 81.3\% \\
 & VQDM & 481 & 497 & 19 & 3 & 48.4\% &  & 479 & 29 & 21 & 471 & \textbf{95.0\%} &  & 472 & 50 & 28 & 450 & 92.2\% \\
 \midrule
Avg Acc &  &  &  &  &  & 50.3\% &  &  &  &  &  & 70.4\% &  &  &  &  &  & \textbf{95.1\%} \\
\bottomrule
\end{tabular}%
\caption{Results for distinguishing belonging images and images generated by other models.}
\label{table-1}
\end{table*}

\begin{table*}[h]
\centering
\small
\setlength{\tabcolsep}{1mm}
\begin{tabular}{ccccccccccccccccccc}
\toprule
\multirow{2.5}{*}{Model} &  & \multicolumn{5}{c}{RONAN} &  & \multicolumn{5}{c}{LatentTracer} &  & \multicolumn{5}{c}{AEDR (ours)} \\
\cmidrule(r){3-7} \cmidrule(lr){9-13} \cmidrule(l){15-19}
 &  & TP & FP & FN & TN & Acc &  & TP & FP & FN & TN & Acc &  & TP & FP & FN & TN & Acc \\
\midrule
SD1.5 &  & 476 & 481 & 24 & 19 & 49.5\% &  & 475 & 83 & 25 & 417 & 89.2\% &  & 493 & 6 & 7 & 494 & \textbf{98.7\%} \\
SD2.1 &  & 477 & 480 & 23 & 20 & 49.7\% &  & 476 & 417 & 24 & 83 & 55.9\% &  & 499 & 1 & 1 & 499 & \textbf{99.8\%} \\
SD2base &  & 475 & 469 & 25 & 31 & 50.6\% &  & 480 & 29 & 20 & 471 & 95.1\% &  & 499 & 2 & 1 & 498 & \textbf{99.7\%} \\
SD3.5 &  & 479 & 393 & 21 & 107 & 58.6\% &  & 478 & 499 & 22 & 1 & 47.9\% &  & 490 & 15 & 10 & 485 & \textbf{97.5\%} \\
SDXL &  & 474 & 479 & 26 & 21 & 49.5\% &  & 478 & 418 & 22 & 82 & 56.0\% &  & 500 & 3 & 0 & 497 & \textbf{99.7\%} \\
FLUX &  & 474 & 370 & 26 & 130 & 60.4\% &  & 478 & 477 & 22 & 23 & 50.1\% &  & 495 & 25 & 5 & 475 & \textbf{97.0\%} \\
VQDM &  & 476 & 473 & 24 & 27 & 50.3\% &  & 477 & 476 & 23 & 24 & 50.1\% &  & 469 & 44 & 31 & 456 & \textbf{92.5\%} \\
KD2.1 &  & 481 & 493 & 19 & 7 & 48.8\% &  & 479 & 83 & 21 & 417 & 89.6\% &  & 472 & 72 & 28 & 428 & \textbf{90.0\%} \\
\midrule
Avg Acc &  &  &  &  &  & 52.2\% &  &  &  &  &  & 66.7\% &  &  &  &  &  & \textbf{96.9\%} \\
\bottomrule
\end{tabular}
\caption{Results for distinguishing belongings and real images. Real images are randomly selected from LAION-5B.}
\label{table-2}
\end{table*}



\section{Experiments and Results}
\label{section 5}

We begin by describing the experimental setup. Next, we assess the effectiveness and efficiency of AEDR in comparison to existing reconstruction-based passive attribution methods, specifically RONAN~\cite{RONAN} and LatentTracer~\cite{LatentTracer}. We then evaluate the generalization capability of AEDR across various autoencoder architectures. Finally, we perform ablation studies to examine the contribution of each component within AEDR.

\subsection{Experimental Setup}
\label{Experiment-1}

\textbf{Models.} We evaluate AEDR on eight text-to-image Latent Diffusion Models, including five Stable Diffusion variants: SD1.5, SD2base, SD2.1, SDXL, and the latest SD3.5. We also include FLUX.1-dev (FLUX), which uses rectified flows. All models use Variational Autoencoders (VAE), except for two based on quantized autoencoders: Kandinsky 2.1 (KD2.1) with Modulating Quantized Vectors, and VQdiffusion (VQDM) using Vector Quantized VAE.

\textbf{Dataset.} We create a dataset of 9,000 images, comprising 1,000 real images sampled from LAION-5B~\cite{LAION} and 8,000 AI-generated images. To minimize image diversity impact, we use the CLIP Interrogator~\cite{CLIP} to extract prompts from the real images. These prompts are then used to generate images with all eight models. The first 500 generated images are for hyperparameter tuning and threshold setting, while the remaining 500 are for performance evaluation. Further details are in Section 4 of the Technical Appendix.

\subsection{Effectiveness}
\label{Experiment-2}

\textbf{Distinguishing Belonging Images from Images Generated by Other Models.}  We compare our method against existing reconstruction-based passive attribution approaches, RONAN~\cite{RONAN} and LatentTracer~\cite{LatentTracer}, to assess efficiency. All methods are tested on the same evaluation dataset across 8 models. For fair comparison, all baselines use official open source implementations.

The experimental results are shown in Table \ref{table-1}, where \( \mathcal{M}_1 \) represents the target model and \( \mathcal{M}_2 \) denotes other models. Our method achieves an average attribution accuracy of 95.1\%, substantially outperforming RONAN and LatentTracer, which achieve 50.3\% and 70.4\%, respectively. While LatentTracer attains over 90\% accuracy on models such as SD1.5, SD2base, and KD2.1, its performance degrades significantly on the remaining five models. The underlying reason is that when the target model is SD1.5, SD2base, and KD2.1, the reconstruction loss is on the order of \(10^{-4}\), whereas for stronger models such as FLUX, the reconstruction loss decreases to approximately \(10^{-5}\) (see Figure~\ref{fig-gr-re} and Sections 1 to 3 of the Technical Appendix). And LatentTracer lacks any mechanism to calibrate the intrinsic complexity of images, resulting in substantial overlap between the distributions of belonging and non‑belonging images, and thus a high false positive rate (FPR). These findings highlight the robustness of AEDR, which improves attribution accuracy by 24.7\% over the strongest baseline.

\textbf{Distinguishing Belonging Images from Real Images.} We further assess the ability of AEDR to distinguish belonging images from real images. As shown in Table~\ref{table-2}, AEDR achieves an average accuracy of 96.9\%, significantly outperforming RONAN (52.2\%) and LatentTracer (66.7\%). These results confirm that AEDR offers a substantial advantage—30.2\% improvement over the best baseline. Notably, both baselines exhibit high FPR, whereas AEDR maintains a low FPR alongside its superior accuracy.

\begin{table}[b]
\centering
\small
\setlength{\tabcolsep}{1mm}
{%
\begin{tabular}{ccccc}
\toprule
Model & SD1.5 & SD2.1 & SD2base & SD3.5 \\
\midrule
LatentTracer & 29.85s & 92.09s & 32.89s & 29.85s \\
AEDR(ours) & 0.27s & 0.62s & 0.25s & 0.69s \\
Speedup & \textbf{110.6}$\times$ & \textbf{148.5}$\times$ & \textbf{131.6}$\times$ & \textbf{43.3}$\times$ \\
\toprule
Model & SDXL & FLUX & VQDM & KD2.1 \\
\midrule
LatentTracer & 162.94s & 30.64s & 12.58s & 48.14s \\
AEDR(ours) & 1.25s & 0.68s & 0.06s & 0.39s \\
Speedup & \textbf{130.4}$\times$ & \textbf{45.1}$\times$ & \textbf{209.7}$\times$ & \textbf{123.4}$\times$ \\
\bottomrule
\end{tabular}%
}
\caption{Running time on different models.}
\label{table-3}
\end{table}

\subsection{Efficiency}
\label{Efficiency}
In this section, we evaluate the computational efficiency of our method. We test all 8 models with 1,000 images each, recording the average runtime for comparison. Due to memory constraints, we use \texttt{bf16} precision for FLUX and SD3.5, while the remaining six models are run with \texttt{fp32} precision. LatentTracer initializes gradient optimization using encoded representations of the input images, while RONAN uses random initialization and thus requires more optimization steps. Hence, our comparison focuses on LatentTracer as the more efficient baseline. As shown in Table~\ref{table-3}, AEDR achieves an approximately \textbf{100$\times$} speedup in runtime efficiency by replacing iterative gradient-based optimization with two simple forward passes through the autoencoder.

\subsection{Generalization}
\label{Experiment-4}
We evaluate the generalization capability of AEDR across three distinct autoencoder architectures: VAE, VQ-VAE, and MoVQ. Table~\ref{table-AE} shows that AEDR achieves an attribution accuracy exceeding 96\% on VAE-based models. On VQ-VAE, the accuracy reaches 90.85\%, and on MoVQ, it drops to 82.93\%. The performance degradation primarily stems from the discrete nature of the latent space in quantized models. Specifically, the input tensor is mapped to the nearest latent code during quantization, introducing an error that hampers precise reconstruction and, consequently, fine-grained attribution. Despite this, AEDR outperforms LatentTracer on VQ-VAE. MoVQ remains a more challenging case, pointing to a promising direction for future research.

\begin{table}[htbp]
    \centering
    \small
    \setlength{\tabcolsep}{1mm}
    \begin{tabular}{ccccc}
    \toprule
    AE Type & Model & RONAN & LatentTracer & AEDR \\
    \midrule
    VAE & SD1.5 & 49.44\% & 91.39\% & \textbf{98.45\%} \\
    VAE & SD2.1 & 49.86\% & 61.70\% & \textbf{99.66\%} \\
    VAE & SD2base & 52.71\% & 96.43\% & \textbf{99.10\%} \\
    VAE & SD3.5 & 49.98\% & 48.96\% & \textbf{96.70\%} \\
    VAE & SDXL & 51.48\% & 63.98\% & \textbf{99.50\%} \\
    VAE & FLUX & 52.26\% & 52.10\% & \textbf{96.65\%} \\
    VQ-VAE & VQDM & 49.86\% & 54.44\% & \textbf{90.85\%} \\
    MoVQ & KD2.1 & 48.88\% & \textbf{93.13\%} & 82.93\% \\
    \bottomrule
    \end{tabular}
    \caption{Generalization to different types of AE.}
    \label{table-AE}
\end{table}

\subsection{Ablation Studies}
\label{Experiment-3}
\textbf{Impact of Reconstruction Loss Metric.} We evaluate the effect of different reconstruction loss metrics on attribution accuracy using the Stable Diffusion v2-1. We utilized 500 belonging images and 500 randomly selected non-belonging images, including both generated and real images. Four commonly used metrics are compared: Mean Absolute Error (MAE), Mean Squared Error (MSE), Structural Similarity Index (SSIM), and Learned Perceptual Image Patch Similarity (LPIPS). As shown in Table~\ref{table-6}, the attribution accuracies are 97.20\% (MAE), 99.40\% (MSE), 99.10\% (SSIM), and 90.70\% (LPIPS), respectively. Therefore, we adopt MSE as the default reconstruction loss metric in AEDR.

\begin{table}[htbp]
\centering
\small
\setlength{\tabcolsep}{1mm}
{%
\begin{tabular}{cccccc}
    \toprule
    Metric & TP & FP & FN & TN & Acc \\
    \midrule
    MAE & 497 & 25 & 3 & 475 & 97.20\% \\
    MSE & \textbf{496} & \textbf{2} & \textbf{4} & \textbf{498} & \textbf{99.40\%} \\
    SSIM & 495 & 4 & 5 & 496 & 99.10\% \\
    LPIPS & 494 & 87 & 6 & 413 & 90.70\% \\
    \bottomrule
\end{tabular}%
}
\caption{Results on different loss metrics.}
\label{table-6}
\end{table}

\textbf{Impact of Homogeneity Calibration.} This section investigates the effect of homogeneity calibration in mitigating the influence of inherent image complexity on attribution accuracy. As reported in Table~\ref{table-5}, homogeneity calibration leads to notable performance gains for most models, with improvements ranging from 0.18\% to 9.09\%. However, FLUX and VQDM~\cite{MoVQ} exhibit slight decreases of 0.15\% and 1.95\%. Despite minor performance drops in a few cases, the overall advantage of the calibration mechanism clearly outweighs these limitations.

\begin{table}[htbp]
\centering
\small
\setlength{\tabcolsep}{1mm}
\begin{tabular}{cccc}
    \toprule
    Model & w/o Calibration & w/ Calibration & Improvement \\
    \midrule
    SD1.5   & 96.29\%          & \textbf{98.45\%} & +2.16\% \\
    SD2.1   & 90.57\%          & \textbf{99.66\%} & +9.09\% \\
    SD2base & 93.71\%          & \textbf{99.10\%} & +5.39\% \\
    SD3.5   & 94.97\%          & \textbf{96.70\%} & +1.73\% \\
    SDXL    & 99.32\%          & \textbf{99.50\%} & +0.18\% \\
    FLUX    & \textbf{96.80\%} & 96.65\%          & -0.15\% \\
    VQDM    & \textbf{92.80\%} & 90.85\%          & -1.95\% \\
    KD2.1   & 78.35\%          & \textbf{82.93\%} & +4.58\% \\
    \bottomrule
\end{tabular}
\caption{Effects of homogeneity calibration.}
\label{table-5}
\end{table}

\textbf{Impact of Quantile Selection.} This section investigates how different quantile selections affect attribution accuracy. Due to the significant variation in the distribution of attribution signals across models, we select model-specific quantile parameters. Each model is tested with 1,000 belonging images: 500 are used for threshold estimation, and the remaining 500 for evaluation. For each model, we select the value of $\alpha$ that yields the highest attribution accuracy on the estimation image set. These selected $\alpha$ values are summarized in Table~\ref{table-4}; further details in Technical Appendix Section 5.

\begin{table}[htbp]
\centering
\small
\setlength{\tabcolsep}{1mm}
{%
\begin{tabular}{ccccc}
\toprule
Model & Alpha & Estimation & Evaluation \\
\midrule
SD1.5 & 0.015 & 98.67\% & 98.48\% \\
SD2.1 & 0.003 & 99.73\% & 99.66\% \\
SD2base & 0.005 & 99.20\% & 98.90\% \\
SD3.5 & 0.035 & 96.15\% & 96.70\% \\
SDXL & 0.003 & 99.29\% & 99.44\% \\
FLUX & 0.02 & 96.28\% & 96.65\% \\
VQDM & 0.085 & 90.47\% & 90.61\% \\
KD2.1 & 0.05 & 83.47\% & 82.93\% \\
\bottomrule
\end{tabular}%
}
\caption{The Selection of quantile selection.}
\label{table-4}
\end{table}

\section{Conclusion}
This paper addresses the critical challenges faced by current reconstruction-based attribution methods when dealing with state-of-the-art (SOTA) generative models. We propose AEDR (AutoEncoder Double Reconstruction), a novel, training-free passive attribution framework for AI-generated images. AEDR integrates a double-reconstruction scheme using an autoencoder with an effective calibration mechanism based on image homogeneity. By computing the reconstruction loss ratio over two consecutive reconstructions, AEDR efficiently mitigates biases arising from inherent image complexity, resulting in significantly improved attribution performance. Extensive experiments demonstrate that AEDR delivers a 25.5\% relative improvement in attribution accuracy while achieving an approximately 100$\times$ speedup in inference time compared to leading baseline methods.

\section*{Acknowledgements}
This work was supported in part by the National Natural Science Foundation of China under Grant 62472398, U2336206, 62121002, 62402469 and 62302146, and by the Fundamental Research Funds of a State Key Laboratory under CBCT2024KF01.

\bibliography{aaai2026}

@inproceedings{SD,
  title={High-resolution image synthesis with latent diffusion models},
  author={Rombach, Robin and Blattmann, Andreas and Lorenz, Dominik and Esser, Patrick and Ommer, Bj{\"o}rn},
  booktitle={Proceedings of the IEEE/CVF conference on computer vision and pattern recognition},
  pages={10684--10695},
  year={2022}
}

@article{GAN,
  title={Generative adversarial nets},
  author={Goodfellow, Ian J and Pouget-Abadie, Jean and Mirza, Mehdi and Xu, Bing and Warde-Farley, David and Ozair, Sherjil and Courville, Aaron and Bengio, Yoshua},
  journal={Advances in neural information processing systems},
  volume={27},
  year={2014}
}

@inproceedings{VAE,
  author       = {Diederik P. Kingma and
                  Max Welling},
  editor       = {Yoshua Bengio and
                  Yann LeCun},
  title        = {Auto-Encoding Variational Bayes},
  booktitle    = {2nd International Conference on Learning Representations, {ICLR} 2014,
                  Banff, AB, Canada, April 14-16, 2014, Conference Track Proceedings},
  year         = {2014},
}

@article{KD2.1,
  title={Kandinsky: an improved text-to-image synthesis with image prior and latent diffusion},
  author={Razzhigaev, Anton and Shakhmatov, Arseniy and Maltseva, Anastasia and Arkhipkin, Vladimir and Pavlov, Igor and Ryabov, Ilya and Kuts, Angelina and Panchenko, Alexander and Kuznetsov, Andrey and Dimitrov, Denis},
  journal={arXiv preprint arXiv:2310.03502},
  year={2023}
}

@article{VQDM,
  title={Improved vector quantized diffusion models},
  author={Tang, Zhicong and Gu, Shuyang and Bao, Jianmin and Chen, Dong and Wen, Fang},
  journal={arXiv preprint arXiv:2205.16007},
  year={2022}
}

@inproceedings{carlini2023extracting,
  title={Extracting training data from diffusion models},
  author={Carlini, Nicolas and Hayes, Jamie and Nasr, Milad and Jagielski, Matthew and Sehwag, Vikash and Tramer, Florian and Balle, Borja and Ippolito, Daphne and Wallace, Eric},
  booktitle={32nd USENIX Security Symposium (USENIX Security 23)},
  pages={5253--5270},
  year={2023}
}

@inproceedings{zhao2021multi,
  title={Multi-attentional deepfake detection},
  author={Zhao, Hanqing and Zhou, Wenbo and Chen, Dongdong and Wei, Tianyi and Zhang, Weiming and Yu, Nenghai},
  booktitle={Proceedings of the IEEE/CVF conference on computer vision and pattern recognition},
  pages={2185--2194},
  year={2021}
}

@inproceedings{chen2020gan,
  title={Gan-leaks: A taxonomy of membership inference attacks against generative models},
  author={Chen, Dingfan and Yu, Ning and Zhang, Yang and Fritz, Mario},
  booktitle={Proceedings of the 2020 ACM SIGSAC conference on computer and communications security},
  pages={343--362},
  year={2020}
}

@inproceedings{ong2021protecting,
  title={Protecting intellectual property of generative adversarial networks from ambiguity attacks},
  author={Ong, Ding Sheng and Chan, Chee Seng and Ng, Kam Woh and Fan, Lixin and Yang, Qiang},
  booktitle={Proceedings of the IEEE/CVF Conference on Computer Vision and Pattern Recognition},
  pages={3630--3639},
  year={2021}
}

@inproceedings{LatentTracer,
  title={How to Trace Latent Generative Model Generated Images without Artificial Watermark?},
  author={Wang, Zhenting and Sehwag Vikash, Chen, Chen and Lyu, Lingjuan and Metaxas, Dimitris N and Ma, Shiqing},
  booktitle={International Conference on Machine Learning},
  year={2024}
}

@inproceedings{RONAN,
  title={Where Did I Come From? Origin Attribution of AI-Generated Images},
  author={Wang, Zhenting and Chen, Chen and Zeng, Yi and Lyu, Lingjuan and Ma, Shiqing},
  booktitle={Thirty-seventh Conference on Neural Information Processing Systems},
  year={2023}
}

@inproceedings{AEroblade,
  title={Aeroblade: Training-free detection of latent diffusion images using autoencoder reconstruction error},
  author={Ricker, Jonas and Lukovnikov, Denis and Fischer, Asja},
  booktitle={Proceedings of the IEEE/CVF Conference on Computer Vision and Pattern Recognition},
  pages={9130--9140},
  year={2024}
}

@article{HFI,
  title={HFI: A unified framework for training-free detection and implicit watermarking of latent diffusion model generated images},
  author={Choi, Sungik and Park, Sungwoo and Lee, Jaehoon and Kim, Seunghyun and Choi, Stanley Jungkyu and Lee, Moontae},
  journal={arXiv preprint arXiv:2412.20704},
  year={2024}
}

@inproceedings{wat-1,
  title={Transparent robust image watermarking},
  author={Swanson, Mitchell D and Zhu, Bin and Tewfik, Ahmed H},
  booktitle={Proceedings of 3rd IEEE International Conference on Image Processing},
  volume={3},
  pages={211--214},
  year={1996},
  organization={IEEE}
}

@article{wat-2,
  title={Reversible image watermarking using interpolation technique},
  author={Luo, Lixin and Chen, Zhenyong and Chen, Ming and Zeng, Xiao and Xiong, Zhang},
  journal={IEEE Transactions on information forensics and security},
  volume={5},
  number={1},
  pages={187--193},
  year={2009},
  publisher={IEEE}
}

@inproceedings{finger-1,
  title={Attributing fake images to gans: Learning and analyzing gan fingerprints},
  author={Yu, Ning and Davis, Larry S and Fritz, Mario},
  booktitle={Proceedings of the IEEE/CVF international conference on computer vision},
  pages={7556--7566},
  year={2019}
}

@inproceedings{finger-2,
  title={Does a GAN leave distinct model-specific fingerprints?},
  author={Ding, Yuzhen and Thakur, Nupur and Li, Baoxin},
  booktitle={BMVC},
  pages={22},
  year={2021}
}

@inproceedings{SD3.5,
  title={Scaling rectified flow transformers for high-resolution image synthesis},
  author={Esser, Patrick and Kulal, Sumith and Blattmann, Andreas and Entezari, Rahim and M{\"u}ller, Jonas and Saini, Harry and Levi, Yam and Lorenz, Dominik and Sauer, Axel and Boesel, Frederic and others},
  booktitle={Forty-first international conference on machine learning},
  year={2024}
}

@article{DDPM,
  title={Denoising diffusion probabilistic models},
  author={Ho, Jonathan and Jain, Ajay and Abbeel, Pieter},
  journal={Advances in neural information processing systems},
  volume={33},
  pages={6840--6851},
  year={2020}
}

@article{DALL,
  title={Improving image generation with better captions},
  author={Betker, James and Goh, Gabriel and Jing, Li and Brooks, Tim and Wang, Jianfeng and Li, Linjie and Ouyang, Long and Zhuang, Juntang and Lee, Joyce and Guo, Yufei and others},
  journal={Computer Science. https://cdn. openai. com/papers/dall-e-3. pdf},
  volume={2},
  number={3},
  pages={8},
  year={2023}
}

@inproceedings{finger-3,
  title={Artificial fingerprinting for generative models: Rooting deepfake attribution in training data},
  author={Yu, Ning and Skripniuk, Vladislav and Abdelnabi, Sahar and Fritz, Mario},
  booktitle={Proceedings of the IEEE/CVF International conference on computer vision},
  pages={14448--14457},
  year={2021}
}

@article{finger-4,
  title={Responsible disclosure of generative models using scalable fingerprinting},
  author={Yu, Ning and Skripniuk, Vladislav and Chen, Dingfan and Davis, Larry and Fritz, Mario},
  journal={arXiv preprint arXiv:2012.08726},
  year={2020}
}

@article{IP-problem,
  title={Towards reliable verification of unauthorized data usage in personalized text-to-image diffusion models},
  author={Li, Boheng and Wei, Yanhao and Fu, Yankai and Wang, Zhenting and Li, Yiming and Zhang, Jie and Wang, Run and Zhang, Tianwei},
  journal={arXiv preprint arXiv:2410.10437},
  year={2024}
}

@article{SDXL,
  title={Sdxl: Improving latent diffusion models for high-resolution image synthesis},
  author={Podell, Dustin and English, Zion and Lacey, Kyle and Blattmann, Andreas and Dockhorn, Tim and M{\"u}ller, Jonas and Penna, Joe and Rombach, Robin},
  journal={arXiv preprint arXiv:2307.01952},
  year={2023}
}

@article{chen2023pathway,
  title={A pathway towards responsible ai generated content},
  author={Chen, Chen and Fu, Jie and Lyu, Lingjuan},
  journal={arXiv preprint arXiv:2303.01325},
  year={2023}
}

@inproceedings{fre-4,
  title={Leveraging frequency analysis for deep fake image recognition},
  author={Frank, Joel and Eisenhofer, Thorsten and Sch{\"o}nherr, Lea and Fischer, Asja and Kolossa, Dorothea and Holz, Thorsten},
  booktitle={International conference on machine learning},
  pages={3247--3258},
  year={2020},
  organization={PMLR}
}

@inproceedings{fre-5,
  title={Think twice before detecting gan-generated fake images from their spectral domain imprints},
  author={Dong, Chengdong and Kumar, Ajay and Liu, Eryun},
  booktitle={Proceedings of the IEEE/CVF conference on computer vision and pattern recognition},
  pages={7865--7874},
  year={2022}
}

@inproceedings{fre-7,
  title={Intriguing properties of synthetic images: from generative adversarial networks to diffusion models},
  author={Corvi, Riccardo and Cozzolino, Davide and Poggi, Giovanni and Nagano, Koki and Verdoliva, Luisa},
  booktitle={Proceedings of the IEEE/CVF conference on computer vision and pattern recognition},
  pages={973--982},
  year={2023}
}

@inproceedings{tex-2,
  title={Detecting AI Generated Images Through Texture and Frequency Analysis of Patches},
  author={Chen, Yuming and Yashtini, Maryam},
  booktitle={2024 4th International Conference on Artificial Intelligence, Virtual Reality and Visualization},
  pages={103--110},
  year={2024},
  organization={IEEE}
}

@inproceedings{tex-4,
  title={Texturecrop: Enhancing synthetic image detection through texture-based cropping},
  author={Konstantinidou, Despina and Koutlis, Christos and Papadopoulos, Symeon},
  booktitle={Proceedings of the Winter Conference on Applications of Computer Vision},
  pages={1459--1468},
  year={2025}
}

@article{tex-7,
  title={Optimized Frequency Collaborative Strategy Drives AI Image Detection},
  author={Li, Jun and Jiang, Wentao and Shen, Liyan and Ren, Yawei},
  journal={IEEE Internet of Things Journal},
  year={2025},
  publisher={IEEE}
}

@article{LAION,
  title={Laion-5b: An open large-scale dataset for training next generation image-text models},
  author={Schuhmann, Christoph and Beaumont, Romain and Vencu, Richard and Gordon, Cade and Wightman, Ross and Cherti, Mehdi and Coombes, Theo and Katta, Aarush and Mullis, Clayton and Wortsman, Mitchell and others},
  journal={Advances in neural information processing systems},
  volume={35},
  pages={25278--25294},
  year={2022}
}

@article{VQ-VAE,
  title={Neural discrete representation learning},
  author={Van Den Oord, Aaron and Vinyals, Oriol and others},
  journal={Advances in neural information processing systems},
  volume={30},
  year={2017}
}

@article{MoVQ,
  title={Movq: Modulating quantized vectors for high-fidelity image generation},
  author={Zheng, Chuanxia and Vuong, Tung-Long and Cai, Jianfei and Phung, Dinh},
  journal={Advances in Neural Information Processing Systems},
  volume={35},
  pages={23412--23425},
  year={2022}
}

@article{KDE,
  title={Robust kernel density estimation},
  author={Kim, JooSeuk and Scott, Clayton D},
  journal={The Journal of Machine Learning Research},
  volume={13},
  number={1},
  pages={2529--2565},
  year={2012},
  publisher={JMLR. org}
}

@article{KDEevaluation,
  title={Evaluation of threshold selection methods for adaptive kernel density estimation in disease mapping},
  author={Ruckthongsook, Warangkana and Tiwari, Chetan and Oppong, Joseph R and Natesan, Prathiba},
  journal={International Journal of Health Geographics},
  volume={17},
  pages={1--13},
  year={2018},
  publisher={Springer}
}

@inproceedings{Gaussian-shading,
  title={Gaussian shading: Provable performance-lossless image watermarking for diffusion models},
  author={Yang, Zijin and Zeng, Kai and Chen, Kejiang and Fang, Han and Zhang, Weiming and Yu, Nenghai},
  booktitle={Proceedings of the IEEE/CVF Conference on Computer Vision and Pattern Recognition},
  pages={12162--12171},
  year={2024}
}

@article{Gaussian-shading++,
  title={Gaussian Shading++: Rethinking the Realistic Deployment Challenge of Performance-Lossless Image Watermark for Diffusion Models},
  author={Yang, Zijin and Zhang, Xin and Chen, Kejiang and Zeng, Kai and Yao, Qiyi and Fang, Han and Zhang, Weiming and Yu, Nenghai},
  journal={arXiv preprint arXiv:2504.15026},
  year={2025}
}

@article{Tree-Ring,
  title={Tree-ring watermarks: Fingerprints for diffusion images that are invisible and robust},
  author={Wen, Yuxin and Kirchenbauer, John and Geiping, Jonas and Goldstein, Tom},
  journal={arXiv preprint arXiv:2305.20030},
  year={2023}
}

@article{use-2,
  title={Diffusion models beat gans on image synthesis},
  author={Dhariwal, Prafulla and Nichol, Alexander},
  journal={Advances in neural information processing systems},
  volume={34},
  pages={8780--8794},
  year={2021}
}

@article{FLUX,
  title={FLUX. 1 Kontext: Flow Matching for In-Context Image Generation and Editing in Latent Space},
  author={Batifol, Stephen and Blattmann, Andreas and Boesel, Frederic and Consul, Saksham and Diagne, Cyril and Dockhorn, Tim and English, Jack and English, Zion and Esser, Patrick and Kulal, Sumith and others},
  journal={arXiv e-prints},
  pages={arXiv--2506},
  year={2025}
}

@misc{CLIP,
  author={Pharmapsychotic},
  year={2023},
  title={Clip-interrogator},
  howpublished = {\url{https://github.com/pharmapsychotic/clip-interrogator}},
  note = {Accessed: 2025-11-26}
}

@inproceedings{van2016pixel,
  title={Pixel recurrent neural networks},
  author={Van Den Oord, A{\"a}ron and Kalchbrenner, Nal and Kavukcuoglu, Koray},
  booktitle={International conference on machine learning},
  pages={1747--1756},
  year={2016},
  organization={PMLR}
}
\end{document}